\documentclass[sn-basic]{sn-jnl}

\jyear{2022}%
\usepackage{indentfirst}
\theoremstyle{thmstyleone}%
\theoremstyle{thmstyletwo}%
\theoremstyle{thmstylethree}%

\usepackage{arydshln}

\usepackage{bbding}
\begin{document}

\title[\tiny Faster Learning of Temporal Action Proposal via Sparse Multilevel Boundary Generator]{Faster Learning of Temporal Action Proposal via Sparse Multilevel Boundary Generator}


\author[1]{\fnm{Qing} \sur{Song}}\email{priv@bupt.edu.cn}

\author[1]{\fnm{Yang} \sur{Zhou}}\email{zhouyang2020@bupt.edu.cn}

\author[1]{\fnm{Mengjie} \sur{Hu}}\email{mengjie.hu@bupt.edu.cn}

\author*[1]{\fnm{Chun} \sur{Liu}}\email{chun.liu@bupt.edu.cn}



\affil[1]{\orgdiv{Pattern Recognition and Intelligent Vision}, \orgname{Beijing University of Posts and Telecommunications}, \orgaddress{\street{Xi Tu Cheng Road}, \city{Beijing}, \postcode{100876},  \country{People's Republic of China}}}



\abstract{Temporal action localization in videos presents significant challenges in the field of computer vision. While the boundary-sensitive method has been widely adopted, its limitations include incomplete use of intermediate and global information, as well as an inefficient proposal feature generator. To address these challenges, we propose a novel framework, Sparse Multilevel Boundary Generator (SMBG), which enhances the boundary-sensitive method with boundary classification and action completeness regression. SMBG features a multi-level boundary module that enables faster processing by gathering boundary information at different lengths. Additionally, we introduce a sparse extraction confidence head that distinguishes information inside and outside the action, further optimizing the proposal feature generator. To improve the synergy between multiple branches and balance positive and negative samples, we propose a global guidance loss. Our method is evaluated on two popular benchmarks, ActivityNet-1.3 and THUMOS14, and is shown to achieve state-of-the-art performance, with a better inference speed (2.47xBSN++, 2.12xDBG). These results demonstrate that SMBG provides a more efficient and simple solution for generating temporal action proposals. Our proposed framework has the potential to advance the field of computer vision and enhance the accuracy and speed of temporal action localization in video analysis.The code and models are made available at \url{https://github.com/zhouyang-001/SMBG-for-temporal-action-proposal}.}


%




\maketitle
\section{Introduction}\label{sec1}


In this section, we provide a comprehensive overview of the current state of video temporal action localization task, and identify areas that require further improvement. Additionally, we conduct a detailed analysis of the limitations of current approaches in this field. Building on these insights, we introduce various optimizations that we have developed on top of boundary-sensitive methods, and present the results of our experiments in this area.

With the proliferation of video content on the internet, the analysis of such content has become an increasingly important area of research in both academia and industry. Of particular interest is the task of temporal action detection, which involves the localization of action instances within long videos, with both action categories and temporal boundaries. This task can be subdivided into two steps: temporal action proposal and action recognition. Although the accuracy of temporal action proposals is generally low, many methods have been developed to address this challenge.

One of the leading approaches is the boundary-sensitive method, as proposed in \cite{2019BMN}. This method utilizes boundary classification and action completeness regression to generate proposals. Despite its potential, several problems still need to be addressed in order to improve the performance of this method.

Temporal action detection is a critical area of research due to its broad range of applications, including video surveillance, sports analysis, and video summarization. As such, continued efforts are necessary to refine and improve existing methods to achieve even greater levels of accuracy and efficiency. The boundary-sensitive method has shown great promise in this regard, and further research is needed to fully explore its capabilities and address its limitations.

Despite the impressive performance of state-of-the-art methods such as BMN \cite{2019BMN} and BSN++ \cite{2020BSN}, there remain two key challenges that need to be addressed in order to further enhance the accuracy and efficiency of these approaches.

\begin{figure}[h]%
	\centering
	\includegraphics[width=0.8\textwidth]{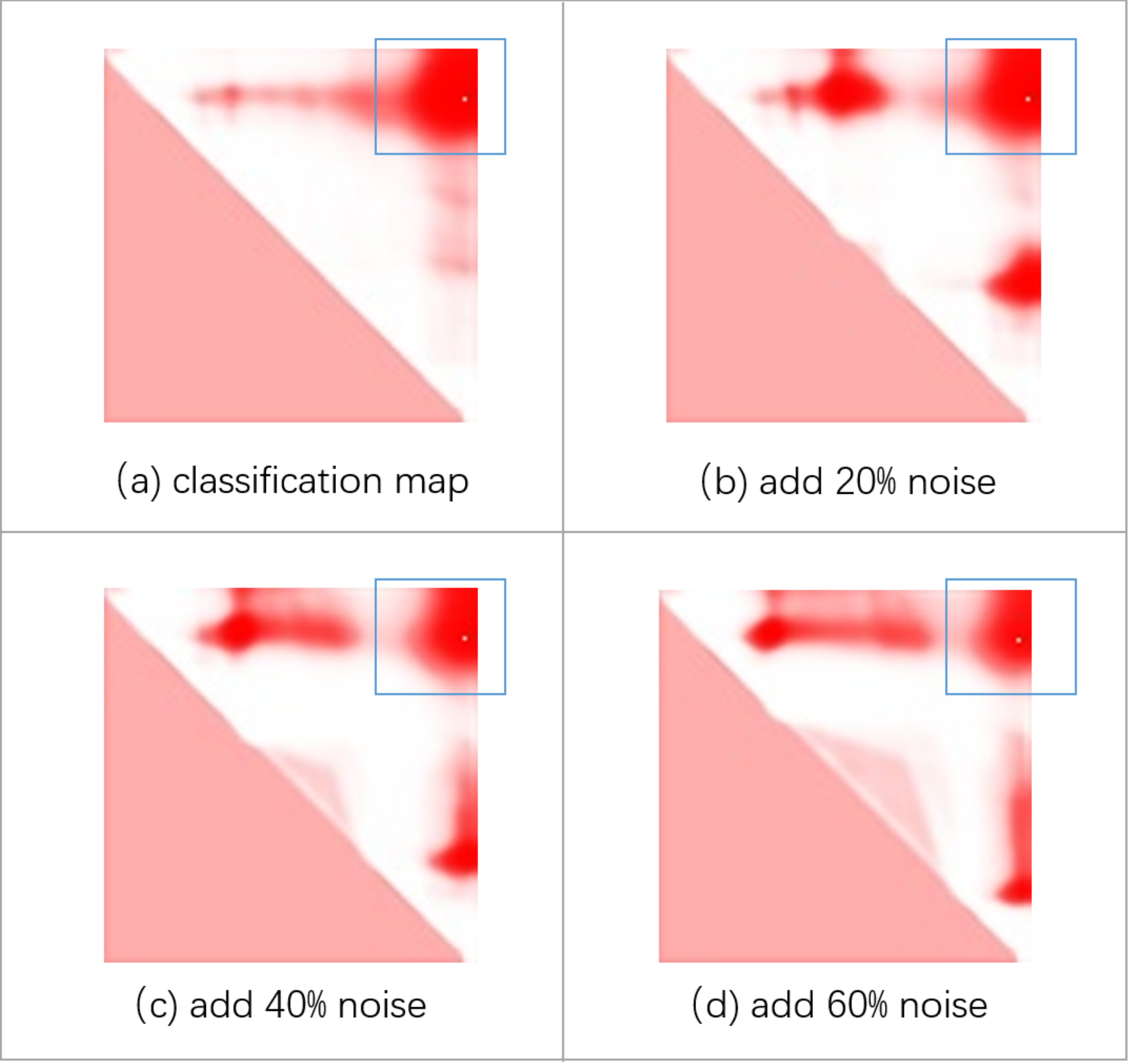}
	\caption{In this image, we use gaussian noise replace the central area of action's feature. We test 20\% , 40\% and 60\% respectively, and show the results of confidence maps. }\label{fig1}
\end{figure}

Firstly, current methods are characterized by insufficient utilization of intermediate and global information. Although some approaches employ proposal feature generators to obtain the middle part of the action's feature, they do not make optimal use of this information. As illustrated in Fig. \ref{fig1}, when a large proportion of the action center's feature is replaced with noise, the action completeness regression exhibits only minor changes, and even recognizes some noise areas as action, indicating a lack of effective discrimination of the central region. Moreover, noise is a common occurrence in video content, and the network should also have the ability to observe the overall situation. For instance, long actions generally do not occur around multiple short actions, and long actions and internal short actions are mutually exclusive.

The second challenge is the efficiency of inference, owing to the complexity of the proposal feature generator. While some recent works have attempted to address this issue by employing the transformer framework \cite{trans} to speed up the process \cite{2021TCA,2021RTD}, or optimizing NMS and the amount of computation of the action completeness region \cite{2020DBG}, there remains a pressing need to optimize the proposal generator itself, rather than merely addressing its symptoms. Notably, some works have also abandoned the action completeness region branch \cite{2020Accurate,2020BC} to expedite training and inference speed.

In conclusion, while the boundary-sensitive method holds great potential in temporal action detection, its further development requires a focus on addressing these key challenges. We believe that optimizing the proposal generator, and more effectively utilizing intermediate and global information, represent promising avenues for future research.

To explore the effectiveness of the feature extraction layer in the boundary sensitive network, we conducted an experiment by simplifying the module to a fully connected (FC) layer. Despite the poor performance of the FC layer in comparison to the traditional boundary sensitive network, which highlights the significant impact of the prior knowledge embedded in the feature extraction layer, we found that the model could still learn certain patterns through training. These observations suggest that providing some prior knowledge to the module and optimizing the complex sampling and 3D convolution processes can lead to improved performance.

Given that convolution is capable of extracting features on a large scale, we considered two 1D convolution-based methods to reduce the computational expansion caused by 2D and 3D processes. The first method involved convolution on the center, with kernel sizes based on the length of the proposal. However, it was challenging to set convolution kernels at different positions due to the significant variation in action length. A higher number of scales would lead to a complex network, slower inference speed, and worse outcomes.

Therefore, we opted for the second method, which involved discarding the middle feature of the action and performing 1D convolution only at the position close to the start and end of the action. Given that we focused on the change in the start and end position, the scale of the convolution kernel did not need to be too complex but could still achieve satisfactory results. These findings underscore the potential of optimizing the proposal generator itself and enhancing the efficiency of inference for the boundary sensitive method, and suggest that further research in this direction could lead to improved temporal action detection.

In addition to optimizing the feature extraction layer, we also explored alternative methods for information extraction within action instances. Several approaches, such as self-attention and deformable convolution, were investigated. However, our experimental results showed that these complex mechanisms did not yield significant improvements in performance. In contrast, we found that dilated convolution, which involves increasing the spacing between kernel weights, was a more effective option. The relative simplicity of this approach may be an advantage, as it is less prone to overfitting and may generalize better to different datasets.

It is worth noting that the lack of appropriate training methods may have hindered the performance of some of these approaches. Further research may be required to develop new training strategies or modifications to existing ones in order to fully explore the potential of these methods.

In conclusion, to overcome the limitations of existing boundary-sensitive action detection methods, we have proposed the Sparse Multilevel Boundary Generator (SMBG) framework. This framework integrates center and background features to predict the boundary map, and features a novel multilevel proposal feature generation layer to replace the traditional approach, which significantly reduces Flops and enhances inference speed. In addition, we designed a sparse extraction confidence head to capture information inside and beside the action, and added feature analysis of the middle part to the 2D branch. To address the issue of sample balance, we introduced a global guidance loss to improve the sampling strategy of positive and negative samples for the confidence map's loss. Compared to other approaches, which optimize the computing kernel or non-maximum suppression (NMS), we focused on optimizing the feature generation layer to extract information around the action. Our experiments show that SMBG achieves comparable accuracy with state-of-the-art methods on the ActivityNet 1.3 dataset while significantly improving inference speed.

In summary, this paper presents several contributions to the field of temporal action proposal:

(1) We introduce Sparse Multilevel Boundary Generator (SMBG), a faster and more efficient approach for temporal action proposal. SMBG distinguishes between information inside and outside the action, and achieves state-of-the-art performance on the ActivityNet 1.3 dataset.

(2) We propose a new proposal feature generation layer that significantly reduces computational complexity through a simpler extraction method. This layer is designed to enhance the detection of information around the action, while also improving accuracy.

(3) To address the problem of sample balance, we introduce a global guidance loss that improves the sampling strategy of other losses, and helps to balance different branches in the network.

Overall, our proposed approach represents a significant improvement over existing methods, and demonstrates the potential for further progress in the field of temporal action proposal.

The overall framework of this paper is as follows. The background and related work are introduced in Section \ref{sec2}. In Section \ref{sec3}, the proposed Sparse Multilevel Boundary Generator is described in detail. In Section \ref{sec4}, the experiments on the ActivityNet-1.3 and THUMOS14 datasets are presented, and the proposed model is compared with the state-of-the-art methods. Finally, the conclusions are drawn in Section \ref{sec5}.

\section{Related Work}\label{sec2}

In this section, we present a detailed overview of the current state of research on video temporal action localization task, and conduct a comprehensive analysis of the strengths and weaknesses of the existing literature in this field.

\textbf{Temporal action proposal.}Temporal action proposal aims to detect action instances with temporal boundaries and confidence in untrimmed videos. Most works can also be divided into two patterns: Anchor-based and Boundary-based. Anchor-based methods generate proposals by designing a set of multi-scale anchors with regular temporal intervals. SCNN \cite{2016Temporal} adopts the C3D network as a binary classifier for anchor evaluation. RGNMF-AN \cite{2022ARGR} make a use of a combination of attributed and topological information in tandem to solve Link prediction. TURN \cite{2017TURN} divide the video into equal length elements and do temporal regression to adjust the action boundaries. IMOPSO \cite{2020AMOPSO} develop a novel feature selection method by integrating of node centrality and PSO algorithm and improves disease diagnosis prediction accuracy.
 
Boundary-based methods evaluate each temporal location in
video. BSN \cite{2018BSN} generate location probabilities to generate temporal boundaries and evaluate proposals' global confidence. While it become a baseline of temporal action proposal,  it also lead to inefficiency because of the repeated confidence calculations. 

To solve this problem, BMN \cite{2019BMN} propose a Boundary-sensitive method, it propose a boundary-matching mechanism for confidence evaluation of proposals in an end-to-end pipeline. Although it works, there is still place for improvement in two aspects.

One is to reduce boundary noise and improve the quality of confidence score, like combines anchor-based method and boundary-based method(MGG \cite{2020MGG}), proposes dual stream BaseNet to generate two different levels and more discriminative features(DBG \cite{2020DBG}), model the insightful relations between the boundary and action content by the graph neural networks(BC-GNN \cite{2020BC}), exploits complementary boundary regressor , relation modeling for temporal proposal generation(BSN++ \cite{2020BSN}) and use auxiliary background constraint idea(BCNet \cite{2022BCNet}). But how to establish the relationship between local and global is still a problem, and lack of attention to the relationship between branches.

The other is to improve efficiency. Some works use Transformer \cite{trans} replace CNN to avoid NMS(TCANet \cite{2021TCA},RTD-Net \cite{2021RTD}), other work optimize the computing kernel of proposal feature generation layer(DBG \cite{2020DBG}). SIBFS \cite{2021ARSI} present a comparative analysis of different feature selection methods. MLCA2F \cite{2022AMLCA2F} use Multi-Level Context Attentional to solve Computed Tomography (CT) images. However, because there is no work to focus on proposal feature generation method, this module still occupies most of the resources and time of inference.

In addition, some work has been improved in other directions, like improve the method of extracting video features from the beginning of feature extraction(ABN \cite{2022ABN}). Other developed a robust CAD system based on transfer learning and multi-layer feature fusion network to diagnose complex skin diseases(\cite{2022ASLR}).

The Sparse Multilevel Boundary Generator (SMBG) proposed in our paper can distinguish information inside and outside the action, and can greatly reduce the calculation of the proposal feature generation layer through a simpler extraction method. We also propose a global guidance loss to assist the balance between different branches and change the sampling strategies in classification loss.

\section{Sparse Multilevel Boundary Generator (SMBG)}\label{sec3}

In this section, we introduce the specific algorithmic improvement, the Sparse Multilevel Boundary Generator, in Section 3.1. We then provide detailed explanations of its training and inference procedures in Section 3.2, which will be presented in subsequent sections.

Our improvements are aimed at proposal feature generation task, so we will first introduce the task.

Suppose there are a set of untrimmed video frames $F = \left \{ f_{t}  \right \} _{t=1 }^{l_{f}}  $, where $f_{t} $ is the $t-\mathrm{t} \mathrm{h} $ RGB frame and $l_{f} $ is the number of frames in the video $V$. The annotation of $V$ can be denoted by a set of action instances $\psi _{g} =\left \{ \varphi _{i} =\left ( t_{si} ,t_{ei}  \right )   \right \}_{i=1}^{N_{g} } $, where $N_{g}$ is the number of ground truth action instances in video $V$, and $t_{si} ,t_{ei}$ are starting and ending points of action instance $\varphi _{i}$. The generation of temporal action proposal aims to predict proposals $\psi _{p}=\left \{ \varphi _{i}=\left ( t_{si},t_{ei},p_{i}    \right )   \right \} _{i=1}^{N_{p} }$ to cover $\psi _{g}$ with high recall and overlap, where $p_{i}$ is the confidence of $\varphi_{i}$.

\subsection{Sparse Multilevel Boundary Generator}\label{sec3.1}

\begin{figure}[h]%
	\centering
	\includegraphics[width=0.9\textwidth]{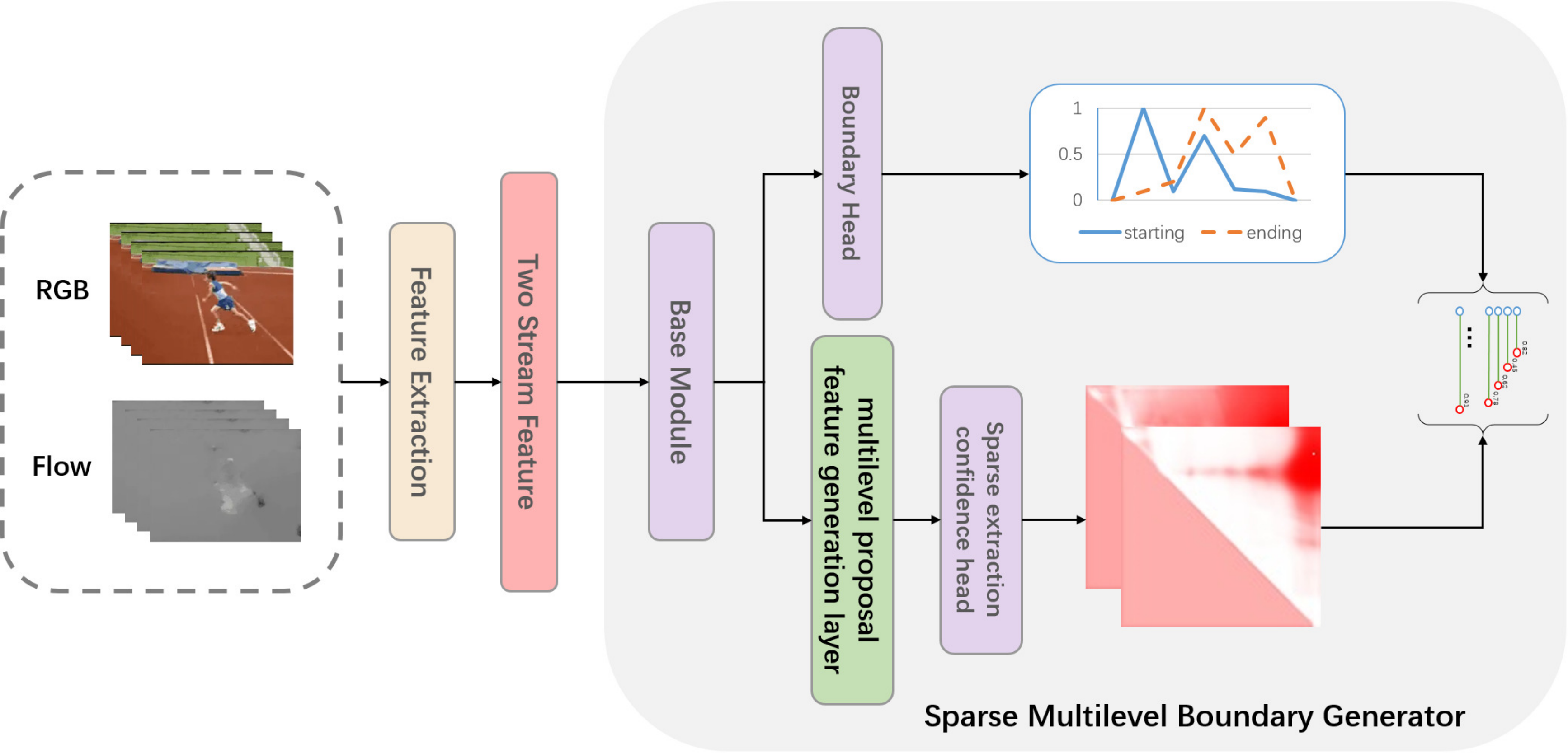}
	\caption{This image shows the structure of our Sparse Multilevel Boundary Generator (SMBG). It contains Base Module, Boundary Head,multilevel proposal feature generation layer and sparse extraction confidence head.}\label{fig2}
\end{figure}

In this section, we present the Sparse Multilevel Boundary Generator (SMBG) as our proposed framework for generating confidence scores for densely distributed proposals. As illustrated in Fig. \ref{fig2}, the spatial-temporal network \cite{2016Temporal} is utilized to encode the visual contents of the video during the video representation phase. The output scores of the two-stream network are used as RGB and flow features, respectively, which are then fed into SMBG. The Base module can be considered as a BaseNet that extracts local behavioral information and is designed to be kept simple to demonstrate the effectiveness of subsequent modules. Subsequently, the Boundary classification module evaluates the starting and ending probabilities for all temporal locations in the untrimmed video. The output of the BaseNet also enters the multilevel proposal feature generation layer, which generates temporal context features for each proposal through multiscale feature merging. The proposal feature is then sent to the sparse extraction confidence head, which produces confidence maps by aggregating sparse global information.

\begin{figure}[h]%
	\centering
	\includegraphics[width=0.9\textwidth]{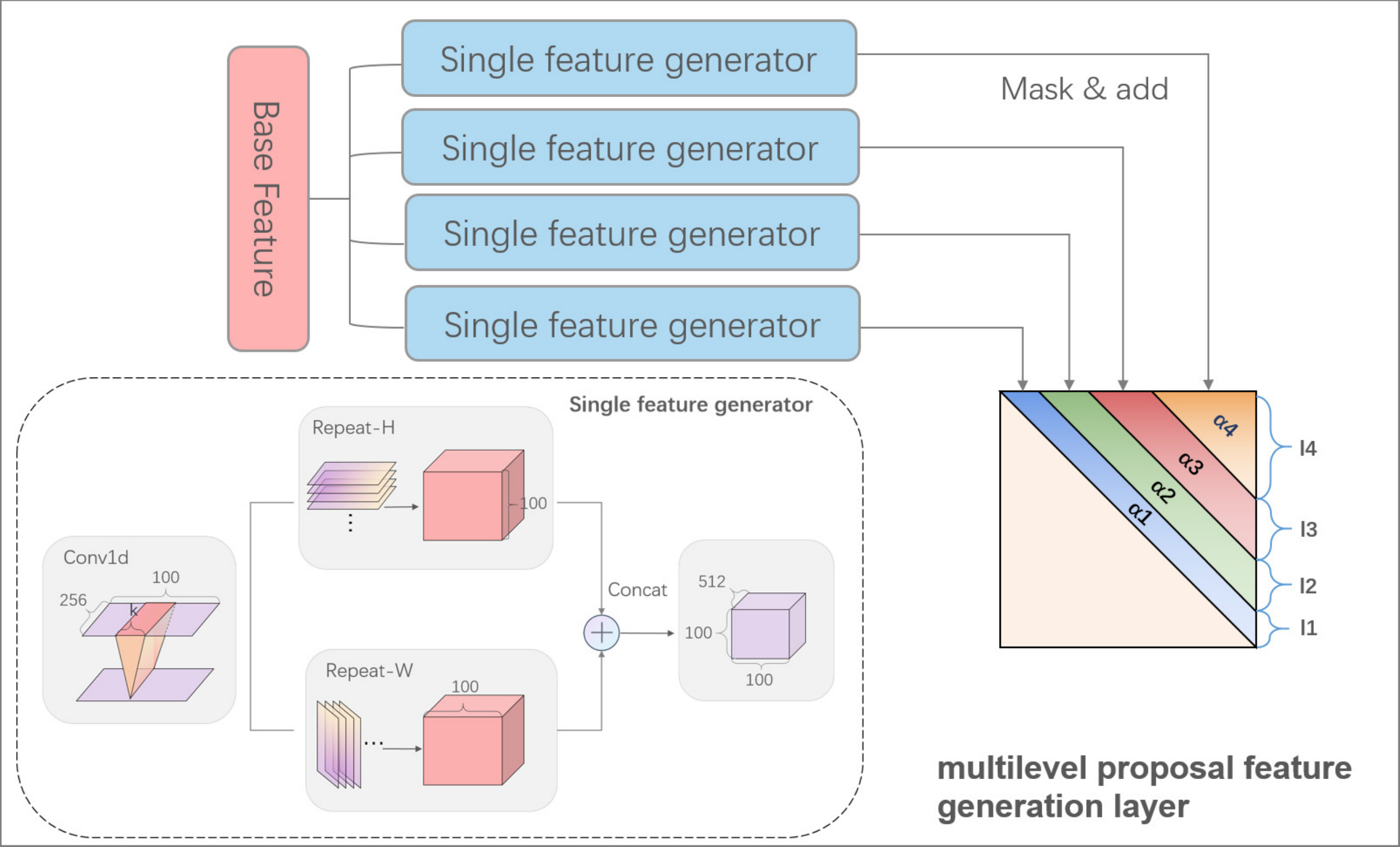}
	\caption{This picture shows the structure of multilevel proposal feature generation layer. The two-stream network's output is sent into four single feature generators. After a 1D convolution, repeat, and concat, the output features are masked on different time lengths $\iota_{1}$, $\iota_{2}$, $\iota_{3}$, and $\iota_{4}$ of the map. Their positive areas are $\alpha_{1}$, $\alpha_{2}$, $\alpha_{3}$, and $\alpha_{4}$. The final proposal feature is the sum of them.}\label{fig3}
\end{figure}

\textbf{Multilevel proposal feature generation layer.}We present a novel multilevel proposal feature generation layer that generates temporal context features for each proposal and allows our framework to be end-to-end trainable. The aim of this design is to significantly reduce the complexity and calculation of the model. In contrast to traditional Boundary-based methods, we focus only on the area around the action boundary, which is divided into four scales, as shown in Fig. \ref{fig3}. The process begins with using 1D convolution of different kernel sizes $k_{1},k_{2},k_{3},k_{4}$ to generate several features at each start and end position. We then repeat these features in the start and end dimensions, respectively, and concatenate them to create several feature maps with different fields. Finally, we mask and sum the feature map at different positions to obtain the final proposal feature $f_{p}$. This process can be expressed mathematically as:

\begin{equation}
	f_{p}= \sum_{i=1}^{4} \alpha _{i} \times Cat\left ( Repeat_{w}\left ( Conv1d\left ( f_{b}  \right )  \right ) ,Repeat_{H}\left ( Conv1d\left ( f_{b}  \right ) \right )  \right ) 
\end{equation}

Where $f_{b} \in R^{B\times N\times T} $ is the output of BaseNet, $B$ is the batch size, $N$ is the depth of feature and $T$ is lenth of time dimention. $\alpha _{i}\in R^{T\times T} $ is the regional weight of $i_{th} $ feature generator, calculate by $\alpha _{i}=\left \{ (x_{m},y_{n} )=1 \text{ if } l_{i-1}<=m<=n<l_{i} \text{ else }0  \right \}$, $l_{i}$ is the parameter selected by experience. $f_{p} \in R^{B\times 2N\times T\times T}$ is the output of multilevel proposal feature generation layer.

Our novel multilevel proposal feature generation layer greatly simplifies the calculation process and reduces the complexity of the model while retaining high accuracy.

We introduce a sparse convolution version by employing dilation convolution instead of traditional convolution. This modification results in a reduction in the number of parameters, while maintaining the level of accuracy.

\textbf{Sparse extraction confidence head.} The aim of the sparse extraction confidence head is to accurately generate confidence maps by effectively distinguishing and utilizing both boundary and surrounding information. To achieve this, we employ a dilation convolution to gather information from the surrounding areas, which is a key difference from a simple confidence head. The complete process is as follows:

\begin{equation}
	P^{c}= F_{sigmoid}  \left ( F_{\left ( Conv1,Conv2,Conv3 \right ) } \left ( F_{DilConv\left ( f_{p}  \right ) }  \right )  \right ) 
\end{equation}

Most layer has a Relu and a BatchNorm layer inside, the dilation rate is $r_{d}$, $P^{c} \in R^{B\times T\times T}$ is the final boundary confidence map, while $B$ is the batch size and $T$ is lenth of time dimention.

\subsection{Training and inference}\label{sec3.2}

To facilitate joint learning using both boundary classification scores and confidence maps, we propose a unified multi-task loss. In the inference phase, the predicted classification scores and confidence maps are combined using Soft-NMS, a variant of Non-Maximum Suppression (NMS), to generate the final set of dense proposals with confidence.

\textbf{Label Assignment.}Our label generation module basicly follow BMN. Specifically, we compose boundary label $g_{s},g_{e}$ for boundary classification loss, and $g_{c}$ for confidence classification loss and regression loss.

Set $t_{s},t{e}$ as the start and end time of a candidate proposal. For a given ground truth action instance $\varphi =\left ( t_{s},t_{e}   \right ) $, define it action region as $r^{c} = \left [ t_{s},t_{e} \right ] $, start region as $r^{s} =\left [ t_{s} -d_{t} ,t_{s} -d_{t}  \right ] $ and end region as $r^{e} =\left [ t_{e} -d_{t} ,t_{e} -d_{t}  \right ] $ where $d_{t}$ is the two temporal locations intervals. $g_{c}$,$g_{s}$ and $g_{e}$ are calculated as maximum overlap ratio IoR of $r^{c}$, $r^{s}$ and $r^{e}$. where IoR is defined as the overlap ratio with ground truth proportional to the duration of this region.

\textbf{Boundary scores' loss.}For each start and end time, given the generated boundary probability sequence $p_{s}$ and $p_{e}$, as well as the boundary label sequence $g_{s}$ and $g_{e}$, we can construct the boundary loss function $L_{B}$ as the sum of starting and ending losses:

\begin{equation}
	L_{B} =L_{bl}\left ( P_{S}, G_{s}  \right ) + L_{bl}\left ( P_{E}, G_{E}  \right )
\end{equation}

Where the $P_{S}, P_{E}, G_{s}, G_{E}$ mean the predicted scores and ground truth of start and end.

The weighted binary logistic regression loss function $L_{bl}\left ( P,G \right ) $ followed \cite{2018BSN}, denoted as:

\begin{equation}
	L_{bl}\left ( P,G \right ) = \frac{1}{l_{\omega } } \sum_{i=1}^{l_{\omega }} \left ( \alpha ^{+  }\cdot b_{i}\cdot \log\left ( p_{i}  \right ) +\alpha ^{-} \cdot \left ( 1-b_{i}  \right ) \cdot \log\left ( 1-p_{i}  \right ) \right )
\end{equation}

Where $b_{i} =sign\left ( g_{i}-\theta   \right ) $ is a two-value function used to convert $g_{i}$ from $\left [ 0,1 \right ] $ to $\left \{ 0,1 \right \} $ based on overlap threshold $\theta =0.5$. Denoting $l^{+} =\sum b_{i} $ and $l^{-} = l_{\omega } -l^{+} $, the weighted terms are $\alpha ^{+} =\frac{l_{\omega } }{l^{+} } $ and $\alpha ^{-} =\frac{l_{\omega } }{l^{-} } $.

\textbf{Confidence maps' loss.}The Confidence maps' loss $L_{C}$ and the set of it's weight is followed by \cite{2019BMN}. which is the sum of binary classification loss and regression loss:

\begin{equation}
L_{C} =L_{c}\left ( P_{c}, G_{C}  \right ) + \lambda  * L_{r}\left ( P_{r}, G_{C}  \right )
\end{equation}

Where we adopt $L_{bl}$ for classification loss $L_{c}$ and $L2$ loss for regression loss $L_{r}$ and set the weight term $\lambda = 10$. $P_{c}, P_{r}$ is the predictive value of classification and regression. $G_{C}$ is the ground truth of Confidence map.

The main difference in our proposed sampling strategy lies in the classification loss, where we utilize a random sampling strategy for negative samples in $L_r$ to balance the proportion of positive and negative samples at 1:5. This modification enables the model to focus more on samples with significant errors, and learn more effectively from misclassified samples.

\textbf{Global guidance loss.} To address the issue of unstable test accuracy resulting from discordance among branches after adding M-PFG, we calculate the comprehensive loss as the global guidance loss. The global guidance loss is computed based on the scoring method used in the final evaluation, so that the branches can be aligned. For both ground truth and predicted values, we multiply the starting and ending position scores with each location on the confidence map. We use $L_{bl}$ to generate the global guidance loss $L_{G}$, which is defined as:

\begin{equation}
	L_{G}\left ( P,G \right ) = L_{bl}\left( P_{m},G_{m}	\right ) 
\end{equation}

$P_{m},G_{m}$ can be defined as:

\begin{equation}
\begin{aligned}
	&P_{m}=Repeat_{w}\left( P_{s}\right) \times Repeat_{h}\left( P_{e}\right) \times P_{c} \times P_{r}\\
	&G_{m}=Repeat_{w}\left( G_{s}\right) \times Repeat_{h}\left( G_{e}\right) \times G_{c} \times G_{r}
\end{aligned}
\end{equation}

The weight of global guidance loss $\beta$ is set to 0.2. The total training objective of SMBG is redefined as:

\begin{equation}
	L_{smbg} = L_{B} + L_{C} + \beta \times  L_{G} 
\end{equation}

\section{Experiments}\label{sec4}

 This chapter presents a comprehensive study on the efficacy of SMBG in the task of video temporal action locations. Our experiments involve comparisons with existing works, ablation experiments, and parameter analysis. The results demonstrate the superior performance of SMBG models in this task, showcasing their potential for advancing the field of computer vision.

\subsection{Evaluation Datasets}\label{sec4.1}

ActivityNet-1.3 and THUMOS14 are most widely used in temporal action proposal generation task, so we conduct experiments on these two datasets.

\textbf{ActivityNet-1.3.} It is a large-scale dataset containing 19,994 videos with 200 activity classes for action recognition, temporal proposal generation and detection. Following \cite{2019BMN}, the quantity ratio of training, validation and testing sets satisfies 2:1:1.

\textbf{THUMOS14.} THUMOS-14 contains 200 and 213 untrimmed videos with temporal annotations of 20 action classes in validation and testing sets respectively.The training set of THUMOS14 is the UCF-101, which contains trimmed videos for action recognition task. 

In this section, we compare our method with SOTA methods on both ActivityNet-1.3 and THUMOS14.

\subsection{Implementation Details}\label{sec4.2}

Following \cite{2019BMN}, for feature encoding, we adopt the two-stream network \cite{2014two}. During feature extraction, the interval $\sigma$ is set to 16 and 5 on ActivityNet-1.3 and THUMOS14 respectively. On ActivityNet-1.3, we rescale the feature sequence of input videos to $l_{w}  = 100$ by linear interpolation, and the maximum duration $D$ is also set to 100 to cover all action instances.

For ActivityNet-1.3, we resize video feature sequence by linear interpolation and set $L= 100$. For THUMOS14, we slide the window on video feature sequence with $overlap= 0.5$ and $L=128$. When training, we use Adam for optimization. The batch size is set to 16. Both Parameter $l_{i}$ in two datasets are set to $\left \{ 17,33-17,57-33,100-57 \right \}  $, and the kernel sizes $\left \{ k_{i},i\in 1,2,3,4 \right \}  $ are set to $\left \{ 17,33,57,99 \right \} $.

\subsection{Comparison Experiments}\label{sec4.3}

To evaluate the proposal quality, we adopt different IoU thresholds to calculate the average recall (AR) with the average number of proposals (AN). Following the conventions, A set of IoU thresholds [0.5:0.05:0.95] is used on ActivityNet-1.3, while a set of IoU thresholds [0.5:0.05:1.0] is used on THUMOS14. For ActivityNet-1.3, the area under the AR vs. AN curve (AUC) is also used as the evaluation metric.

\textbf{Comparison to the state-of-the-arts.} Tab. \ref{tab1} illustrates the comparison results on ActivityNet-1.3. It can be observed that SMBG has a very similar accuracy compared with SOTA methods, which means that our method with a very simple structure without any enhancement module like attention, can play a very good effect.

\begin{table}[]
	\begin{center}
		\begin{minipage}{180pt}
			\caption{Comparison between SMBG with other state-of-the-art methods on ActivityNet-1.3 dataset in terms of AR@AN and AUC.}\label{tab1}%
			\begin{tabular}{lll}
				\hline
				Method    & AR@100 (val) & AUC (val) \\ \hline
				Prop-SSAD & 73.01        & 64.40     \\
				CTAP      & 73.17        & 65.72     \\
				BSN       & 74.16        & 66.17     \\
				MGG       & 74.54        & 66.43     \\
				BMN       & 75.01        & 67.10     \\
				BSN++     & 76.52        & \textbf{68.26}     \\
				TCANet    & 76.08        & 68.08     \\
				RTD-Net   & 73.21        & 65.78     \\
				DBG       & \textbf{76.65}        & 68.23     \\ \hline
				SMBG(Ours)      & 75.89        & 68.09     \\ \hline
			\end{tabular}
		\end{minipage}
	\end{center}
\end{table}

Tab. \ref{tab2} compares proposal generation methods on the testing set of THUMOS14. To ensure a fair comparison, we adopt the same video feature and post-processing step. Tab. 3 shows that our method using two-stream video features has a similar performance compared with SOTA when the proposal number is set within [50,100,200,500,1000].

\begin{figure}[h]%
	\centering
	\includegraphics[width=0.9\textwidth]{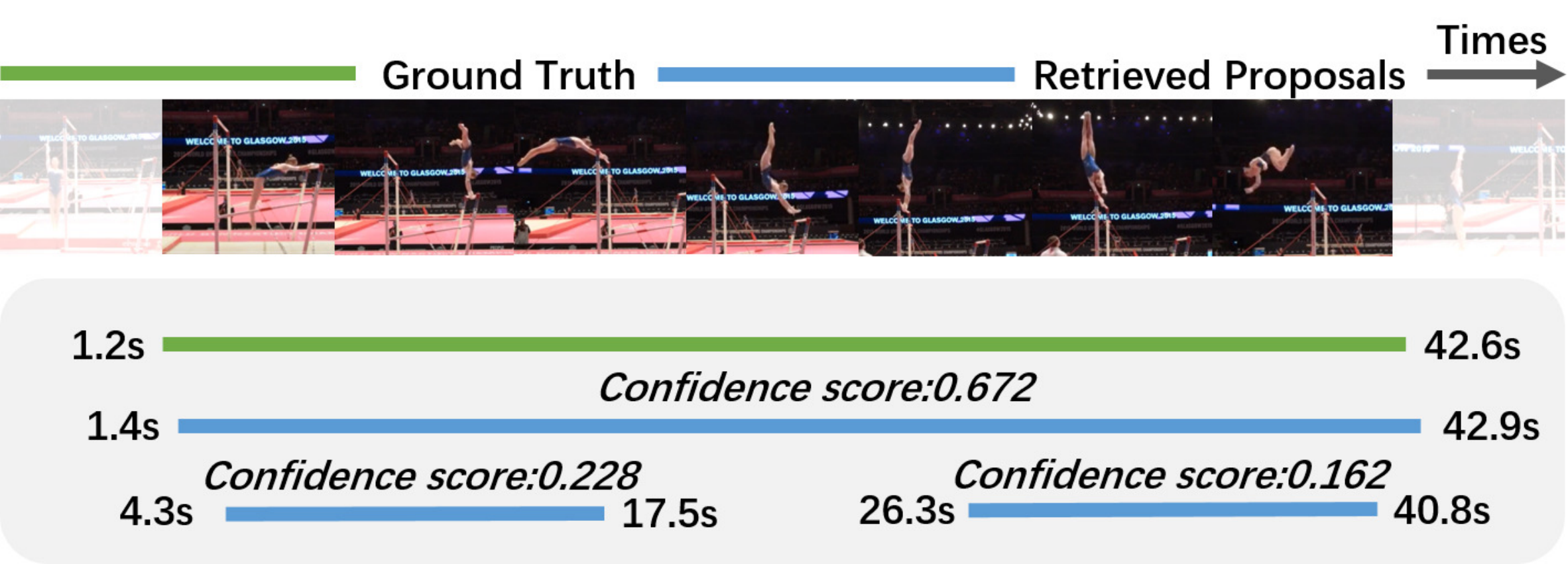}
	\caption{Qualitative examples of proposals generated by SMBG on ActivityNet-1.3. Where each bar represents an action and the start and end times of the actual action are displayed on the left and right, respectively. The green bars indicate the Ground True, while the blue bars represent the predicted proposals with their confidence scores displayed within each box. This visualization enables an intuitive understanding of the accuracy and precision of our proposed approach for temporal action localization in videos.}\label{fig4}
\end{figure}

We show a qualitative example on ActivityNet-1.3 in Fig. \ref{fig4}.

\begin{table}[]
	\begin{center}
		\begin{minipage}{300pt}
			\caption{Comparison between SMBG with other state-of-the-art methods on THUMOS14 in terms of AR@AN.}\label{tab2}%
			\begin{tabular}{lllllll}
				\hline
				Feature  & Method       & @50            & @100           & @200           & @500           & @1000          \\ \hline
				2-Stream & TURN         & 21.86          & 31.89          & 43.02          & 57.63          & 64.17          \\
				2-Stream & MGG          & 39.93          & 47.75          & 54.65          & 61.36          & 64.06          \\
				2-Stream & BSN(SNMS)    & 37.46          & 46.06          & 53.21          & 60.64          & 64.52          \\
				2-Stream & BMN(SNMS)    & 39.36          & 47.72          & 54.70          & 62.07          & 65.49          \\
				2-Stream & BSN++(SNMS)  & \textbf{42.44} & 49.84          & \textbf{57.61} & \textbf{65.17} & 66.83          \\
				2-Stream & DBG(SNMS)    & 37.32          & 46.67          & 54.50          & 62.21          & 66.40          \\
				2-Stream & TCANet(SNMS) & 42.05          & \textbf{50.48} & 57.13          & 63.61          & \textbf{66.88} \\ \hline
				2-Stream & SMBG(SNMS)   & 40.34          & 48.72          & 56.30          & 62.67          & 66.13          \\ \hline
			\end{tabular}
		\end{minipage}
	\end{center}
\end{table}

\textbf{Efficiency study.} We compare our method to other boundary sensitive methods, such as BSN, BMN, BSN++ (reproduced version), and DBG, on the validation set of ActivityNet-1.3. The evaluation focuses on both the effectiveness and efficiency of the models. Our method is tested on a NVIDIA Tesla P40, and the results in Tab. \ref{tab3} demonstrate its superior performance. Notably, for a 3-minute video, our method achieves significantly faster inference speed. We observe that the proposal feature generation layer is the most time-consuming component, and our design effectively addresses this issue.

\begin{table}[h]
	\begin{center}
		\begin{minipage}{200pt}
			\caption{Comparison of Efficiency in ActivityNet-1.3.}\label{tab3}%
			
			\begin{tabular}{lllll}
				\hline
				Method                   & AR@100 & AUC   & $T_{pro}$  & $T_{all}$   \\ \hline
				BSN                      & 74.16  & 66.17 & 475.7 & 478.25 \\
				BMN                      & 75.01  & 67.10 & 25.18 & 27.56  \\
				BSN++ 					 & 76.52  & \textbf{68.26} & 20.54 &  23.13 \\
				DBG                      & \textbf{76.65}  & 68.23 & 16.72 & 19.91  \\ \hline \hline
				SMBG(Ours) 					 & 75.89  & 68.09 & \textbf{6.51} & \textbf{9.36}   \\ \hline
			\end{tabular}
			\footnotetext{Efficiency comparison among BSN, BMN, BSN++, DBG and our method SMBG in validation set of ActivityNet-1.3. We use ms as the time unit. Experiments show the higher speed(3x of BMN, 2x of DBG) of our method which remain high accuracy.} 
		\end{minipage}
	\end{center}
\end{table}

\begin{figure}[h]%
	\centering
	\includegraphics[width=0.8\textwidth]{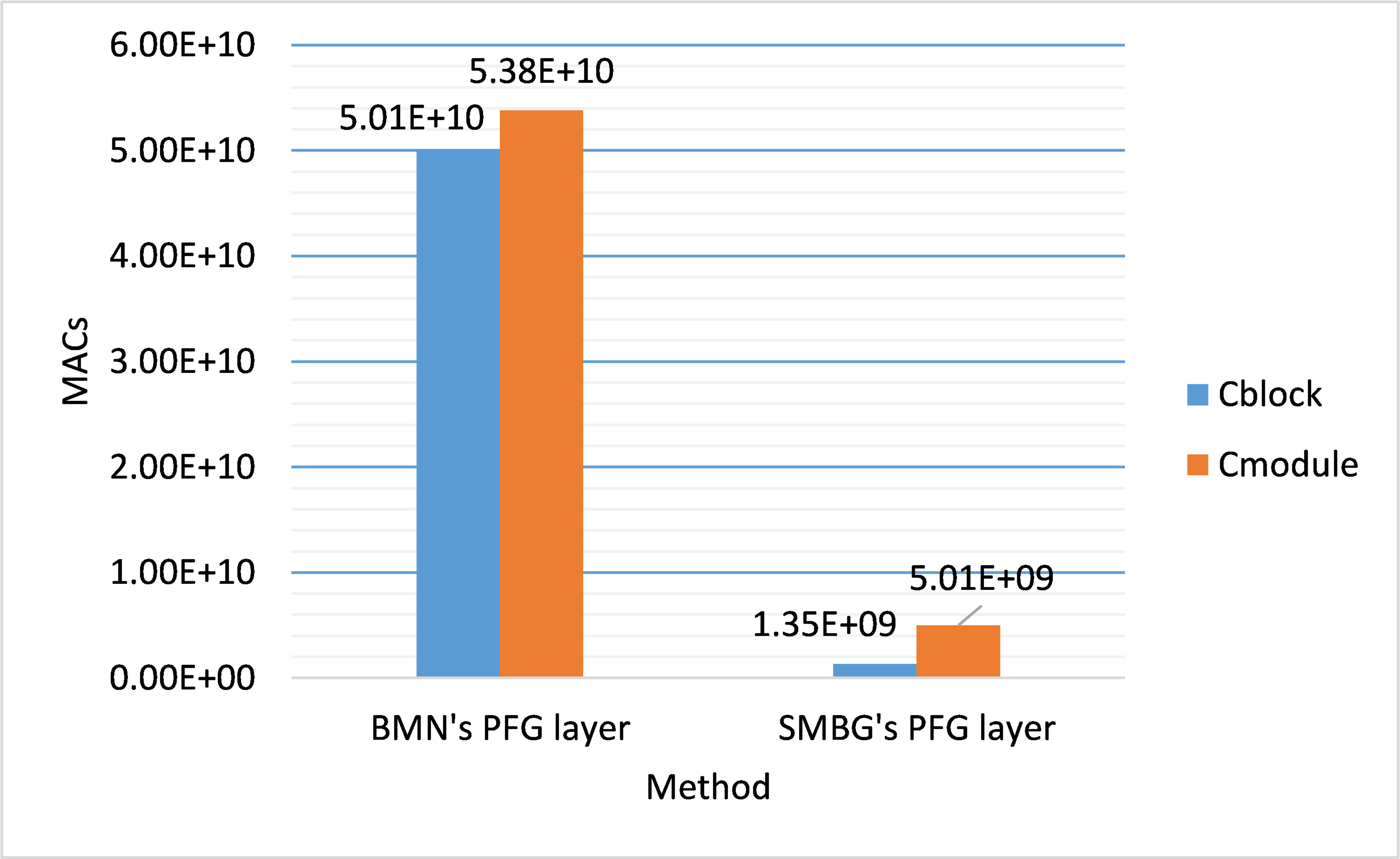}
	\caption{This figure show the calculation amount(in MACs) of different methods. $C_{block}$ is the amount of calculation of different proposal feature generation layers. $C_{module}$ is the calculation modules. Our multilevel proposal feature generation layer reduce the amount of calculation from $5.01\times 10^{10}$MAC to $1.35\times 10^{9}$MAC(2.7\% of BMN). Calculation amount of the whole module is reduced from $5.38\times 10^{10}$MAC to $5.01\times 10^{9}$MAC(9.3\% of BMN).}\label{fig5}
\end{figure}

\textbf{Comparison of calculation amount.} To assess the benefits of our approach, we calculate the amount of computation of our method and BMN by MACs. As shown in Fig. \ref{fig5}, we calculate the proposal feature generation layer's and whole model's amount of calculation. The MACs of BMN's PFG layer accounts for 93.1\% of the whole BMN. Our SMBG's multilevel proposal feature generation layer reduce the MACs from $5.01\times 10^{10}$MAC to $1.35\times 10^{9}$MAC(2.7\% of BMN). Calculation amount of the whole module is reduced from $5.38\times 10^{10}$MAC to $5.01\times 10^{9}$MAC(9.3\% of BMN). These facts show that our method has a huge reduction in the amount of calculation.

\subsection{Ablation study}\label{sec4.4}

This section presents an ablation study on ActivityNet-1.3 to verify the effectiveness of each module in SMBG. The results in Tab.\ref{tab4} show that PFG is the original proposal feature generation layer in BMN, while M-PFG is our multilevel proposal feature generation layer. GG-loss is the global guidance loss, which can solve the branch disharmony problem caused by the simplification of the model. With the GG-loss, our M-PFG outperforms the original PFG. Additionally, the SEC-head, which extracts sparse confidence information, significantly improves the AUC by 0.51, demonstrating the effectiveness of distinguishing internal information for the task. Finally, the different sampling strategies used in our work, denoted as Sampling+, also contribute to the task.

\begin{table}[h]
	\begin{center}
		\begin{minipage}{260pt}
			\caption{Comparison of Calculation}\label{tab4}%
			\begin{tabular}{llllll}
				\hline
				PFG        & M-PFG      & GG-loss    & SEC-head   & Sampling+  & AUC   \\ \hline
				\checkmark &            &            &            &            & 67.10 \\
				& \checkmark &            &            &            & 66.79 \\
				& \checkmark & \checkmark &            &            & 67.34 \\
				& \checkmark & \checkmark & \checkmark &            & 67.85 \\
				& \checkmark & \checkmark & \checkmark & \checkmark & 68.09 \\ \hline
			\end{tabular}
		\end{minipage}
	\end{center}
\end{table}

\textbf{Analysis of generation layer's kernel size.} To evaluate the impact of the M-PFG layer, we conducted experiments to analyze the effect of different kernel sizes on proposal generation performance. Tab.\ref{tab5} shows that a kernel size of 41 for the single feature generator achieved the best performance. Moreover, the four-branches proposal feature generation layer showed the best performance. The kernel size that performed best was consistent with the uniform distribution of samples.

\begin{table}[h]
	\begin{center}
		\begin{minipage}{330pt}
			\caption{Generation layer's kernel size study in ActivityNet-1.3.}\label{tab5}%
			\setlength{\tabcolsep}{3pt}
			\begin{tabular}{l:ccc:cc:cc}
				\hline
				only k or k1/.../kn & 31    & 41             & 51    & 17/41/81       & 21/41/71       & 15/29/53/81    & 17/33/57/99    \\ \hline
				AR@5                & 48.87 & \textbf{49.25} & 48.67 & \textbf{49.30} & 48.76          & 49.18          & \textbf{49.79} \\
				AR@10               & 56.1  & \textbf{56.41} & 56.01 & \textbf{56.86} & 55.88          & 56.79          & \textbf{57.23} \\
				AR@100              & 74.47 & \textbf{74.62} & 74.32 & 74.80          & \textbf{74.94} & \textbf{74.85} & 74.71          \\
				AUC                 & 66.7  & \textbf{66.99} & 66.54 & \textbf{67.21} & 67.06          & 67.19          & \textbf{67.34} \\ \hline
			\end{tabular}
			\footnotetext{This table show the set of different kernel size in M-PFG and their performance(only with M-PFG and GG-loss).} 
		\end{minipage}
	\end{center}
\end{table}

\textbf{Analysis of dilation rate.}This experiment show different dilation rate $r_{d}$ 's influence on model accuracy. As shown in Tab. \ref{tab6}, when the $r_{d}$ is set to 7, the sparse extraction confidence head get the best result.

\begin{table}[h]
	\begin{center}
		\begin{minipage}{160pt}
			\caption{Different dilation rate of sparse extraction confidence head and it's AUC in ActivityNet-1.3.}\label{tab6}%
			\setlength{\tabcolsep}{3pt}
			\begin{tabular}{llllll}
				\hline
				$r_{d}$ & 5                        & 6                         & 7                         & 8     & 9     \\ \hline
				AUC     & \multicolumn{1}{r}{67.5} & \multicolumn{1}{r}{67.65} & \multicolumn{1}{r}{\textbf{67.85}} & 67.70 & 67.57 \\ \hline
			\end{tabular}
		\end{minipage}
	\end{center}
\end{table}

\section{Conclusion}\label{sec5}

We propose Sparse Multilevel Boundary Generator (SMBG), a more efficient framework for generating temporal action proposals in video analysis. SMBG enhances the traditional Boundary-sensitive method with a multi-level boundary module that enables faster processing by gathering boundary information at different lengths. We also introduce a sparse extraction confidence head that distinguishes information inside and outside the action and a global guidance loss that optimizes the balance of positive and negative samples.

Our extensive experiments on ActivityNet-1.3 and THUMOS14 demonstrate that SMBG achieves significant improvement in efficiency while maintaining great performance. Future work can explore the potential of SMBG in other areas of computer vision and refine the proposed framework to further optimize the accuracy and efficiency of temporal action localization in videos. Overall, our approach provides a more efficient and effective solution for generating temporal action proposals, with the potential to advance the field of video analysis.

\section*{Declarations}
This work was supported by the National Key R\&D Program of China (Grant No.2022YFC3302200).

Conflict of Interest : The authors declare that they have no conflict of interest.

Data sharing not applicable to this article as no datasets were generated or analyzed during the current study.

\bibliography{sn-bibliography}

\begin{thebibliography}{20}
\providecommand{\natexlab}[1]{#1}
\providecommand{\url}[1]{{#1}}
\providecommand{\urlprefix}{URL }
\providecommand{\doi}[1]{\url{https://doi.org/#1}}
\providecommand{\eprint}[2][]{\url{#2}}
 \bibcommenthead

\bibitem[{Bai et~al(2020)Bai, Wang, Tong, Yang, Liu, and Liu}]{2020BC}
Bai Y, Wang Y, Tong Y, et~al (2020) Boundary content graph neural network for
  temporal action proposal generation. In: European Conference on Computer
  Vision, Springer, pp 121--137

\bibitem[{Bakkouri and Afdel(2020)}]{2022ASLR}
Bakkouri I, Afdel K (2020) Computer-aided diagnosis (cad) system based on
  multi-layer feature fusion network for skin lesion recognition in dermoscopy
  images. Multimedia Tools and Applications 79(29-30):20,483--20,518

\bibitem[{Bakkouri and Afdel(2022)}]{2022AMLCA2F}
Bakkouri I, Afdel K (2022) Mlca2f: Multi-level context attentional feature
  fusion for covid-19 lesion segmentation from ct scans. Signal, Image and
  Video Processing pp 1--8

\bibitem[{Gao et~al(2017)Gao, Yang, Chen, Sun, and Nevatia}]{2017TURN}
Gao J, Yang Z, Chen K, et~al (2017) Turn tap: Temporal unit regression network
  for temporal action proposals. In: Proceedings of the IEEE international
  conference on computer vision, pp 3628--3636

\bibitem[{Gao et~al(2020)Gao, Shi, Wang, Li, Yuan, Ge, and Zhou}]{2020Accurate}
Gao J, Shi Z, Wang G, et~al (2020) Accurate temporal action proposal generation
  with relation-aware pyramid network. In: Proceedings of the AAAI Conference
  on Artificial Intelligence, pp 10,810--10,817

\bibitem[{Lin et~al(2020)Lin, Li, Wang, Tai, Luo, Cui, Wang, Li, Huang, and
  Ji}]{2020DBG}
Lin C, Li J, Wang Y, et~al (2020) Fast learning of temporal action proposal via
  dense boundary generator. In: Proceedings of the AAAI Conference on
  Artificial Intelligence, pp 11,499--11,506

\bibitem[{Lin et~al(2018)Lin, Zhao, Su, Wang, and Yang}]{2018BSN}
Lin T, Zhao X, Su H, et~al (2018) Bsn: Boundary sensitive network for temporal
  action proposal generation. In: Proceedings of the European conference on
  computer vision (ECCV), pp 3--19

\bibitem[{Lin et~al(2019)Lin, Liu, Li, Ding, and Wen}]{2019BMN}
Lin T, Liu X, Li X, et~al (2019) Bmn: Boundary-matching network for temporal
  action proposal generation. In: Proceedings of the IEEE/CVF International
  Conference on Computer Vision, pp 3889--3898

\bibitem[{Liu et~al(2019)Liu, Ma, Zhang, Liu, and Chang}]{2020MGG}
Liu Y, Ma L, Zhang Y, et~al (2019) Multi-granularity generator for temporal
  action proposal. In: Proceedings of the IEEE/CVF Conference on Computer
  Vision and Pattern Recognition, pp 3604--3613

\bibitem[{Nasiri et~al(2022)Nasiri, Berahmand, and Li}]{2022ARGR}
Nasiri E, Berahmand K, Li Y (2022) Robust graph regularization nonnegative
  matrix factorization for link prediction in attributed networks. Multimedia
  Tools and Applications pp 1--24

\bibitem[{Qing et~al(2021)Qing, Su, Gan, Wang, Wu, Wang, Qiao, Yan, Gao, and
  Sang}]{2021TCA}
Qing Z, Su H, Gan W, et~al (2021) Temporal context aggregation network for
  temporal action proposal refinement. In: Proceedings of the IEEE/CVF
  Conference on Computer Vision and Pattern Recognition, pp 485--494

\bibitem[{Rostami et~al(2020)Rostami, Forouzandeh, Berahmand, and
  Soltani}]{2020AMOPSO}
Rostami M, Forouzandeh S, Berahmand K, et~al (2020) Integration of
  multi-objective pso based feature selection and node centrality for medical
  datasets. Genomics 112(6):4370--4384

\bibitem[{Rostami et~al(2021)Rostami, Berahmand, Nasiri, and
  Forouzandeh}]{2021ARSI}
Rostami M, Berahmand K, Nasiri E, et~al (2021) Review of swarm
  intelligence-based feature selection methods. Engineering Applications of
  Artificial Intelligence 100:104,210

\bibitem[{Shou et~al(2016)Shou, Wang, and Chang}]{2016Temporal}
Shou Z, Wang D, Chang SF (2016) Temporal action localization in untrimmed
  videos via multi-stage cnns. In: Proceedings of the IEEE conference on
  computer vision and pattern recognition, pp 1049--1058

\bibitem[{Simonyan and Zisserman(2014)}]{2014two}
Simonyan K, Zisserman A (2014) Two-stream convolutional networks for action
  recognition in videos. Advances in neural information processing systems 27

\bibitem[{Su et~al(2021)Su, Gan, Wu, Qiao, and Yan}]{2020BSN}
Su H, Gan W, Wu W, et~al (2021) Bsn++: Complementary boundary regressor with
  scale-balanced relation modeling for temporal action proposal generation. In:
  Proceedings of the AAAI Conference on Artificial Intelligence, pp 2602--2610

\bibitem[{Tan et~al(2021)Tan, Tang, Wang, and Wu}]{2021RTD}
Tan J, Tang J, Wang L, et~al (2021) Relaxed transformer decoders for direct
  action proposal generation. In: Proceedings of the IEEE/CVF International
  Conference on Computer Vision, pp 13,526--13,535

\bibitem[{Vaswani et~al(2017)Vaswani, Shazeer, Parmar, Uszkoreit, Jones, Gomez,
  Kaiser, and Polosukhin}]{trans}
Vaswani A, Shazeer N, Parmar N, et~al (2017) Attention is all you need.
  Advances in neural information processing systems 30

\bibitem[{Vo et~al(2021)Vo, Yamazaki, Truong, Tran, Sugimoto, and Le}]{2022ABN}
Vo K, Yamazaki K, Truong S, et~al (2021) Abn: Agent-aware boundary networks for
  temporal action proposal generation. IEEE Access 9:126,431--126,445

\bibitem[{Yang et~al(2022)Yang, Wu, Wang, Jin, Xia, Yao, and Huang}]{2022BCNet}
Yang H, Wu W, Wang L, et~al (2022) Temporal action proposal generation with
  background constraint. In: Proceedings of the AAAI conference on artificial
  intelligence, pp 3054--3062

\end{thebibliography}

\section{Symple Table}

\begin{table}[!ht]
	\centering
	\begin{tabular}{|l|p{8cm}|}
		\hline
		Symbol & Meaning \\ \hline
		$F$ & untrimmed video frames \\ \hline
		$f_{t} $ & $t-\mathrm{t} \mathrm{h} $ RGB frame \\ \hline
		$N_{g}$ &  the number of ground truth action instances \\ \hline
		$t_{si} ,t_{ei}$   & starting and ending points of action instance $\varphi _{i}$ \\ \hline
		$p_{i}$ &  the confidence of $\varphi_{i}$ \\ \hline
		$k_{1},k_{2},k_{3},k_{4}$ & different kernel sizes for 1D conv  \\ \hline
		$f_{p} \in R^{B\times 2N\times T\times T}$ & the output of multilevel proposal feature generation layer  \\ \hline
		$f_{b} \in R^{B\times N\times T} $ &  the output of BaseNet, $B$ is the batch size, $N$ is the depth of feature and $T$ is lenth of time dimention. \\ \hline
		$\alpha _{i}\in R^{T\times T} $ &  the regional weight of $i_{th} $ feature generator \\ \hline
		$r_{d}$ & the dilation rate  \\ \hline
		$P^{c} \in R^{B\times T\times T}$ & the final boundary confidence map, while $B$ is the batch size and $T$ is lenth of time dimention. \\ \hline
		$t_{s},t{e}$ & the start and end time of a candidate proposal \\ \hline
		$d_{t}$ & the two temporal locations intervals \\ \hline
		$L_{B}$ & the boundary loss function  \\ \hline
		$L_{bl}\left ( P,G \right ) $ & The weighted binary logistic regression loss function  \\ \hline
		$L_{c}$ & classification loss  \\ \hline
		$L_r$ & regression loss \\ \hline
		$L_{G}$ & global guidance loss \\ \hline
		$\beta$ & the weight of global guidance loss  \\ \hline
	\end{tabular}
\end{table}

\end{document}